\documentclass{jmlr}

\jmlrproceedings{arXiv}{arXiv preprint}

\usepackage{booktabs}
\usepackage{float}

\hypersetup{
  hypertexnames=false,
  pdftitle={Activation-Space Order-Swap Geometry: A Site-Asymmetry Audit},
  pdfauthor={Anqi Peter Li},
  pdfsubject={Composed activation steering and first-order measurement artifacts}
}

\theorembodyfont{\upshape}
\theoremheaderfont{\scshape}
\theorempostheader{:}
\theoremsep{\newline}
\newtheorem{observation}{Observation}

\jmlrvolume{}
\firstpageno{1}
\jmlryear{2026}
\jmlrworkshop{Symmetry and Geometry in Neural Representations}

\title[Order-Swap Geometry in Activation Steering]{Activation-Space Order-Swap Geometry:
A Site-Asymmetry Audit}

\author{\Name{Anqi Peter Li} \Email{peterli@substrate-labs.org}\\
\addr Substrate Labs}

\newcommand{\Br}{\mathrm{Br}}
\newcommand{\vi}{v_i}
\newcommand{\vj}{v_j}
\newcommand{\lin}{\mathcal{L}}
\newcommand{\selfterm}{\mathcal{S}}
\newcommand{\mixed}{\mathcal{M}}
\newcommand{\diff}{\delta}
\newcommand{\Trait}{\mathcal{T}}
\newcommand{\Dmat}{\mathsf{D}}

\begin{document}
\raggedbottom

\maketitle

\begin{abstract}\noindent
Order-dependent activation statistics are often interpreted as evidence of interaction, but that interpretation can be confounded by where interventions enter the network. We introduce a no-fit site-asymmetry audit. For a twice-differentiable readout, the open-path order-swap decomposes into a canonical additive response measured by single interventions and an antisymmetrized second difference free of first-order and pure self-curvature terms to second order. Across six open-weight language-model families, the single-intervention baseline explains 84.3-97.7 percent of the bracket norm (mean 93.7 percent), while the no-interaction self-curvature term is 1.8-5.2 times larger than the corrected residual in the two families with the plus/minus injection split. The corrected residual clears a generic-interaction null in three of six families under a confound-free prompt split and two of six after configuration robustness. A known-positive surrogate recovers planted mixed interaction, while a matched site-separation test changes the baseline share and a random architecture reproduces the first-order regime. The same estimator transfers to released non-language references: trained residual fractions fall below a fixed Gaussian-direction null in 11/12 contrasts (5/6 ViT-B/16, 6/6 ResNet-50), a portability check rather than pooled evidence. The contribution is a reusable measurement criterion: run the single-intervention baseline before reading an order-swap vector as interaction or geometric structure; if it explains the vector, form the second difference instead. All claims are scoped to activation-space interventions at distinct sites; we do not claim that representation geometry is globally Abelian.
\end{abstract}

\section{Introduction}\label{sec:intro}

\begin{figure}[H]
\centering
\includegraphics[width=\textwidth]{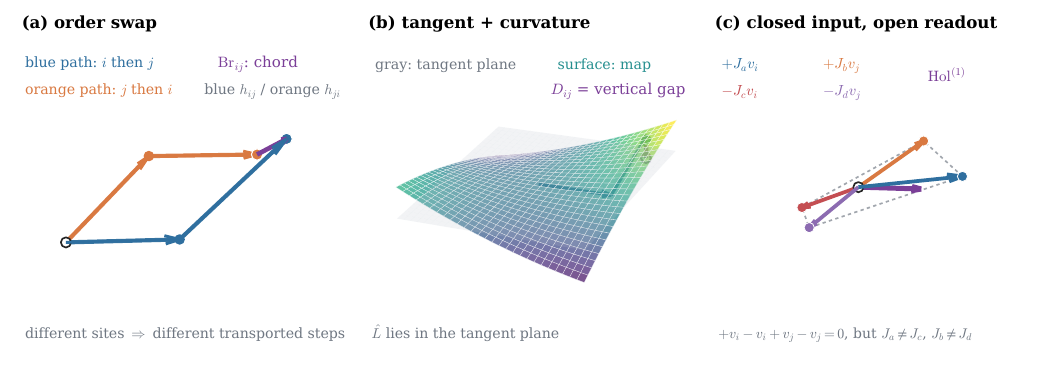}
\caption{\textbf{Three-dimensional schematic of the measurement, not model data.}
\textbf{(a)} Opposite site orders give different Jacobian-weighted paths and an endpoint chord
$\Br_{ij}$. \textbf{(b)} The tangent-plane prediction $\widehat{\lin}$ is separated from the
second-difference curvature $D_{ij}$. \textbf{(c)} Four legs can have zero net injected vector
but a nonzero first-order readout residual when site Jacobians differ. Endpoint order and loop
closure are diagnostics to measure, not evidence of interaction by themselves.}
\label{fig:3d}
\end{figure}

Steering vectors added to a transformer's residual stream compose, and the composition is
order-dependent: injecting direction $\vi$ before $\vj$ does not give the same result as
the reverse. The algebraic reading of that order-dependence is available and inviting. The
measured quantity looks like a bracket: it is antisymmetric, it vanishes when the two
directions coincide, and it is non-zero in practice. In \emph{weight} space the reading is
already established, with the commutator of two fine-tuning updates treated as the governing
order-dependent quantity \citep{sweeney2026geometry}; in activation space the analogous
closed-loop statistic is being computed and read geometrically
\citep{gemma2026holonomy,sevetlidis2026holonomy}. Composable-intervention and task-arithmetic
pipelines provide related composition settings rather than this bracket statistic
\citep{kolbeinsson2025composable,ilharco2023task,obrien2024broadskills}.
Closest to our statistic, \citet{mudarisov2026ffsteering} do form an activation-space Lie
bracket and describe it as ``measuring whether their order
matters''; our criterion returns \emph{exempt} there on the merits rather than by failing to
apply.
We are not aware of a published paper that computes the open-path activation-space
order-swap statistic across \emph{different} injection depths and reads it as evidence of
non-Abelian structure, and we do not attribute that error to anyone. What we supply is a null
model for that statistic and a cheap test of it: a reusable site-asymmetry audit that separates
what the construction forces from what the model contributes.
The nearest published statistic of our own form is in weight space and outside what our
activation-space criterion tests (\citealp{schessl2026pathdependence}; discussed with our own
failed attempt to use it in Appendix~\ref{app:nulls}).
The criterion is also not binary underneath: contamination
is driven by $J_1 - J_2$, so it predicts a gradient, and widening site separation with the
readout held fixed raises the first-order term's share of the bracket in $6$ of $6$ families
and $91.7\%$ of $720$ matched pairs, where raw depth does not and an untrained network of the
same shape reaches only $54$--$58\%$
(Appendix~\ref{app:nulls}; we report this as a count, not a $p$-value, because the six
families share one trait inventory and are not six independent draws). A loop through four
distinct depths is first-order dominated while the degenerate loop is inert
(Figure~\ref{fig:3d}) --- a mechanism-specific test rather than another binary verdict. The audit is organized by a
single principle: \textbf{order-swap geometry has a canonical site-asymmetry baseline that
must be measured before it is interpreted.} Define a quantity by subtracting two orderings and
you get antisymmetry for free, vanishing on repeated arguments for free, and --- when the site
mismatch acts non-trivially --- a measurable baseline for free. The Taylor identity is
elementary; the positive claim is that this is the correct audit object for the class: it is
measured by four single-intervention passes, falsifiable by a matched site-separation
perturbation, and validated by known-positive controls. The six-family result then shows that
the baseline accounts for most of the measured bracket in this setting.

Concretely, the difference expands to first order as $(J_1 - J_2)(\vi - \vj)$, linear in each
direction separately, carrying no interaction, and generically non-zero when the site mismatch
acts on the direction difference. The same pattern recurs at every level we look: the second-order term contains a
self-curvature piece that is also interaction-free and is $1.8$--$5.2\times$ \emph{larger}
than the zero-parameter residual at the primary configuration; closed transport loops, widely assumed exempt, carry
the same first-order term; and the natural diagnostic for conjunctive structure returns
near-zero for \emph{any} antisymmetric form at any rank, so its collapse in real data is
entailed rather than observed. Four apparent findings, one cause.

\paragraph{Contribution.}
\textbf{The central contribution is a site-asymmetry audit for order-dependent activation
statistics, demonstrable without fitting anything.} The
zero-parameter prediction $\widehat{\lin}_{ij} = [s_1(\vi){+}s_2(\vj)] - [s_1(\vj){+}s_2(\vi)]$,
built from four single-injection forward passes and never shown the bracket it predicts,
leaves $5$--$8\%$ of the bracket's norm unexplained at the primary configuration in all six
families, and $2$--$16\%$ across all eighteen family-by-configuration
cells (mean cosine $0.9973$, never below $0.9869$). The cosine and the residual
fraction are one measurement in two units, not two agreeing measurements
(Appendix~\ref{app:jdirect}).

The consequence for practice is a partition, and it is what carries past our own models. Two
statistics are in circulation for ``do these two interventions interact'': the order-swap
bracket $\Br_{ij} = h_{ij} - h_{ji}$, and the \emph{second difference}
$D_{ij} = h_{ij} - s_1(\vi) - s_2(\vj)$. They differ by exactly the artifact above,
$\Br_{ij} = (D_{ij} - D_{ji}) + \widehat{\lin}_{ij}$. The identity is a tautology; the
magnitude is not. Across our eighteen cells the first-order term accounts for
$84.3$--$97.7\%$ of the bracket (mean $93.7\%$), against $0\%$ of the second
difference. Neither half of that contrast is an independent measurement: the first is $1$
minus the residual reported in Section~\ref{sec:results}, the same quantity in complementary
units, and the second is exact by the definition of $D$, surviving even a constant map. This
gives a practical stop condition: run the single-injection prediction before interpreting $\Br$
geometrically; if it explains the bracket, the statistic has not identified interaction. A
reader who wants the interaction should form $D$ rather than correct a bracket to recover it.
This also fixes the status of our own residue, which \emph{is} the
antisymmetrized second difference (Section~\ref{sec:related}).

\begin{figure}[H]
\centering
\includegraphics[width=\textwidth]{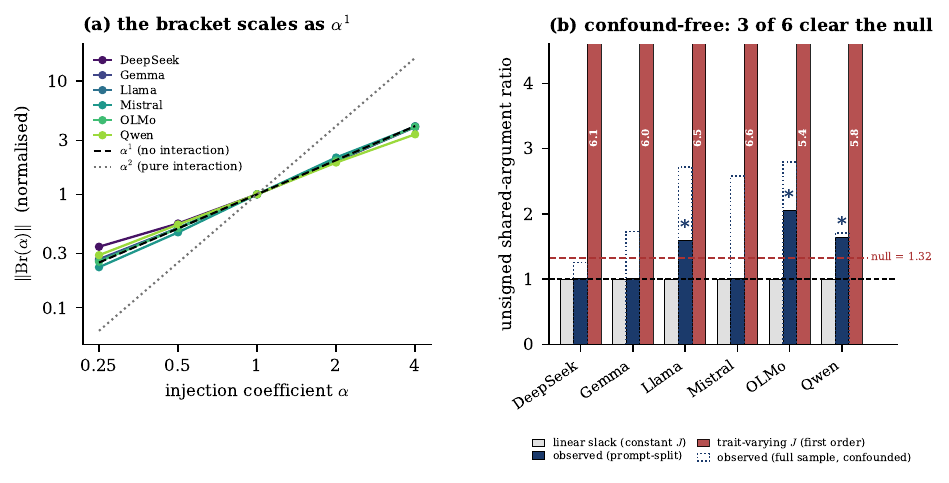}
\caption{\textbf{The two central results.} \textbf{(a)} Norm of the order-swap bracket against
injection coefficient $\alpha$, normalised at $\alpha = 1$. All six families track
$\alpha^{1}$ (dashed), the scaling of a term with no interaction, rather than $\alpha^{2}$
(dotted); fitted exponents $\gamma \in [0.886, 1.043]$. \textbf{(b)} The unsigned
shared-argument ratio for the corrected residual under the \emph{confound-free} prompt-split
design (solid), against the generic-interaction null (dashed red), a calibrated trait-varying
first-order arm, and an uncalibrated constant-Jacobian floor (the linear slack, which has no
free parameter to calibrate; Appendix~\ref{app:nulls}).
Stars mark the $3$ of $6$ families that clear the null --- Llama $1.599$, OLMo $2.059$,
Qwen $1.642$ against the $1.324$ bar; DeepSeek, Gemma and Mistral fall
to linear-slack level. The dotted outline is the same statistic on the full sample, where all
pairs share one prompt set: the gap is what the shared-prompt confound was worth. We plot the
number we claim, not the more favourable full-sample one. The trait-varying arm is clipped at
the axis top and annotated with its true value.}
\label{fig:main}
\end{figure}

\textbf{Second, the correction people would apply does not isolate an interaction either.}
Expanding one order further splits the $\alpha^2$ coefficient in two: a genuinely mixed term
$\mixed$, and a self-curvature term $\selfterm$ that is quadratic in each direction
\emph{separately} and carries no coupling.
We measure $\selfterm$ with no fit. It is $1.8$--$5.2\times$
\emph{larger} than the zero-parameter residual at the primary configuration, so
$\|Q\|/\|\lin\|$ from a two-term fit is not an interaction share: the same artifact
recurring one order up. \citet{pairwisefragile2026} separate these second-order objects at a
\emph{shared} base point; what is new is that when the sites differ, self-curvature
contaminates the order-swap coefficient itself through $H_{11} - H_{22}$.

\emph{Supporting measurements.} Two further routes agree by different apparatus that the
statistic is first-order dominated (Appendix~\ref{app:jdirect}), and a randomly initialised
network reproduces the same agreement, fixing what the six-family measurement establishes:
not that trained models have a property, but that they do not escape a generic one. What
survives correction clears a generic-interaction null in $3$ of $6$ families under the
confound-free design, which we report as a negative result.

\section{Related Work}\label{sec:related}

\citet{vaidyanathan2026curse} prove the second-difference interaction equals the Hessian
bilinear form, vanishing for locally affine maps; that object is our $\mathcal{Q}$, and we
take the theory as given. Note what it entails: a bilinear form applies to
argument pairs it was never fitted on, so transfer to unseen arguments is a property of the
model class, not a discovery. \citet{locallinear2026} and \citet{pairwisefragile2026} both
support the premise that a first-order term dominates here, the first finding layer-wise LLM
dynamics well approximated by locally-linear models, the second finding single perturbations
first-order predictable across nine transformers while pairwise composition has no stable
radius. \citet{steerablereal2026} make the same move for a different statistic: apparent
scale-dependent steerability across 17 models turns out to be produced by an uncalibrated
pipeline; \citet{heap2026randomtransformers} do the same for SAE auto-interpretability with
a randomised baseline, the closest neighbour to our untrained null
(Appendix~\ref{app:nulls}).
The closest methodological neighbour is \citet{attribpatching2026}, who find attribution
patching's first-order approximation unreliable, trace the error to downstream
non-linearity, and supply a correction: the template is ours; the object is not.
\citet{vanderweele2014fourway} provides a causal-inference analogue of the
$\lin + Q$ split, and \citet{bilinearae2026} independently
find low-rank quadratic structure prevalent in LLM activations --- which would be the natural
home for the interaction we are looking for, though we do not find it there: the operator we
recover is not low-rank once the estimator's own rank compression is accounted for
(the full operator comparison is in the artifact).

\paragraph{Weight-space and neighbouring statistics.}
\citet{pairwisefragile2026} measure the analogous Lie bracket for sequential task-gradient
steps, while \citet{sweeney2026geometry} use a weight-space bracket as an ordering statistic;
both are outside our activation-space criterion. The two spaces admit a first-order
correspondence under the conditions analyzed by \citet{adila2026weightact}, so agreement across them is closer to entailment than to an
independent replication. Our untrained null and self-term split are the additional controls.
Among activation-space neighbours, \citet{gemma2026holonomy} use one fixed layer window,
\citet{sevetlidis2026holonomy} prove an input-space affine null, and
\citet{mudarisov2026ffsteering} use a same-site bracket where the first-order term cancels
(the full comparison is in the artifact). \citet{vaidyanathan2026curse} and
\citet{khemais2026crosslayer} instead form the second difference, which is first-order-free.
Composable interventions expose order effects without forming this statistic
(\citet{kolbeinsson2025composable}; task arithmetic \citep{ilharco2023task});
\citet{ortizjimenez2023tangent} study the closest weight-space first-order question.
Exemption is conditional. The second difference cancels first-order content by construction;
a raw bracket at a nominally shared site requires exact $J_1 = J_2$, which is not stable.
Because contamination is first order in the edit while interaction is second order, even a
$1\%$ Jacobian mismatch leaves the former at $99.6\%$ of the bracket at $\alpha = 0.01$.
Closure is not the criterion; site structure is (Section~\ref{sec:estimator}), and the
artifact records the neighbouring statistics treated outside its scope.

\section{Method: the estimator and its first-order term}\label{sec:estimator}

\paragraph{Where the artifact bites.} A measurement is affected when the two interventions
enter at sites with different downstream maps: order-swap brackets, sequential-edit
differences, and any $A\!\to\!B$ versus $B\!\to\!A$ comparison at distinct depths. Closure is
not itself a defence: the first-order terms cancel pairwise only when each return leg
re-enters at its outgoing site, and that composition is the identity map.

Let $A_i^{(\ell)}$ add direction $\vi$ to the residual stream at layer $\ell$, and let
$h^{(\ell_{\mathrm{out}})}$ read out at a later layer. For $\ell_1 < \ell_2$ the order-swap bracket
is $\Br_{ij} = \mathbb{E}_{p}[\, h^{(\ell_{\mathrm{out}})}(A_j^{(\ell_2)} A_i^{(\ell_1)} p)
- h^{(\ell_{\mathrm{out}})}(A_i^{(\ell_2)} A_j^{(\ell_1)} p) ]$.
Expanding about the unsteered activations and carrying the second order out in full,
\begin{equation}
\begin{split}
\Br_{ij} \;=\;& \underbrace{(J_1 - J_2)(\vi - \vj)}_{\lin,\ \text{first order, no interaction}}
\;+\; \underbrace{\tfrac{1}{2}\big[H_{11}(\vi,\vi)-H_{11}(\vj,\vj)
+H_{22}(\vj,\vj)-H_{22}(\vi,\vi)\big]}_{\selfterm,\ \text{second order, \emph{no interaction}}} \\[4pt]
&+\; \underbrace{H_{12}(\vi,\vj) - H_{12}(\vj,\vi)}_{\mixed,\ \text{the interaction}}
\;+\; O(\alpha^3),
\end{split}
\label{eq:expansion}
\end{equation}
where $H_{kk}$ is the curvature of the readout in the perturbation entering at site $k$ and
$H_{12}$ the mixed second derivative.

\begin{observation}[site-asymmetry audit identity]\label{obs:audit}
For matched single-site responses, the canonical additive baseline is
$\widehat{\lin}_{ij}=[s_1(\vi)+s_2(\vj)]-[s_1(\vj)+s_2(\vi)]$, and
$\Br_{ij}-\widehat{\lin}_{ij}=D_{ij}-D_{ji}$. Consequently, to second order the residual
removes both the site-Jacobian and pure self-curvature terms; what remains is the
antisymmetrized mixed derivative plus $O(\alpha^3)$.
\end{observation}

For $C^3$ readouts, Appendix~\ref{app:jdirect} gives an explicit third-derivative norm
certificate for the $O(\alpha^3)$ remainder. We do not claim empirical bounds on those
derivatives for the six LLMs, so their residuals remain finite-scale mixtures unless such
bounds are supplied.

\paragraph{The second-order term is not all interaction.} $\selfterm$ is antisymmetric under
exchange, vanishes when $\vi = \vj$, and is non-zero exactly when the two sites have
different curvature --- every surface property that makes the statistic look like a bracket.
But it is quadratic in each argument \emph{separately}, with no interaction between them: the
first-order artifact one order up. Only $\mixed$ couples the two directions. Writing the
$\alpha^2$ coefficient as a single ``bilinear'' object $\mathcal{Q}(\vi,\vj)$, which is the
natural thing to write, silently merges $\selfterm$ into the interaction and repeats at
second order the exact conflation the paper exists to correct. Two consequences: a two-term
fit $\alpha \lin + \alpha^2 \mathcal{Q}$ estimates $\selfterm + \mixed$ together, so
$\|\mathcal{Q}\|/\|\lin\|$ in the injection-scale fit is \emph{not} an interaction share; and
Observation~\ref{obs:one} guarantees only that the fitted $W(\vi - \vj)$ removes \emph{linear}
structure, which is weaker, because $\selfterm$ is nonlinear too.

\begin{observation}[elementary]\label{obs:one}
$\Br_{ij} = -\Br_{ji}$ by construction. Any linear model $M_1 \vi + M_2 \vj$ that is
antisymmetric under exchange requires $M_1 = -M_2$, hence has the form $W(\vi - \vj)$.
\end{observation}

Fitting and subtracting the best $W(\vi - \vj)$ therefore removes all linear structure in the
model class. This is the elementary symmetric/antisymmetric splitting of the $S_2$ action on
$\Trait \oplus \Trait$, stated because it dictates the correct control rather than as a
result. It is not exact for the \emph{estimate}: with $\widehat{W}$ fitted from finite data
the residual retains $(W - \widehat{W})(\vi - \vj)$, and Section~\ref{sec:results} tests that
this slack does not manufacture our result.

Three natural controls do \emph{not} remove $\lin$. \textbf{(C1)} Projecting off
$\mathrm{span}(\vi, \vj)$: $\lin$ lands outside that span at the readout. \textbf{(C2)} $\lin$ is
antisymmetric, so brackets sharing a direction in \emph{swapped} argument slots are
anti-correlated; a statistic that pools argument positions averages $+$ against $-$.
\textbf{(C3)} $\lin$ alone reproduces which pair a bracket came from, so pair-identifiability
is no evidence of interaction either.

\paragraph{Measuring the first-order term instead of fitting it.}\label{par:jdirect}
Both arguments above are indirect, and the fitted correction of Observation~\ref{obs:one}
carries $16 \times D_{\mathrm{out}}$ free parameters, so a high explained variance is partly a
statement about capacity. We therefore measure $J_1$ and $J_2$ rather than fitting them. The
single-injection response $s_k(v) = \mathbb{E}_p[h^{(\ell_{\mathrm{out}})}(A_v^{(\ell_k)} p) -
h^{(\ell_{\mathrm{out}})}(p)] = J_k v + \tfrac{1}{2}H_{kk}(v,v) + O(\alpha^3)$ costs one forward
pass, and $\widehat{\lin}_{ij} = [s_1(\vi) + s_2(\vj)] - [s_1(\vj) + s_2(\vi)]$
predicts the bracket with \emph{no fitted parameters}; it never sees $\Br_{ij}$. Expanding to
second order, every Jacobian and every pure self-term cancels, leaving
$\Br_{ij} - \widehat{\lin}_{ij} = H_{12}(\vi,\vj) - H_{12}(\vj,\vi) + O(\alpha^3)$: the
antisymmetrized \emph{mixed} second derivative at second order, plus an uncontrolled
higher-order remainder at finite injection scale. Thus this route targets the same interaction
term as the fitted correction, but its finite-scale residual is not an exact interaction estimate.
The estimator is falsifiable: the
symmetric combination should not predict an antisymmetric object, and a bracket paired with a
\emph{different} pair's responses should not be predicted at all. (Exchanging the two site
labels negates $\widehat{\lin}$ identically, so that check tests our arithmetic, not the
model.) Appendix~\ref{app:jdirect} validates it on
surrogates with known $J_1, J_2, H_{11}, H_{22}, H_{12}$.

\paragraph{Known-positive control.} On known-Hessian surrogates, the no-interaction arm
recovers the bracket exactly; a pure mixed arm with $J_1=J_2$ gives $\widehat{\lin}=0$ and
residual fraction $1$; and mixed arms with self-curvature recover the planted interaction share
to machine precision at three strengths (Appendix~\ref{app:jdirect}). The model-side residual
remains a finite-scale mixture, not an interaction estimate.

\paragraph{Closed loops: site structure, not closure.}
A holonomy statistic is often assumed exempt because its net injection is zero. That argument
is wrong: it uses two Jacobians for four injections, so the loop is the identity map and the
statistic is identically zero, leaving nothing to exempt. A genuine loop through four distinct
depths gives $(J_1 - J_3)\vi + (J_2 - J_4)\vj$, which does not vanish and carries no
interaction (the loop checks are in the artifact). Exemption is earned from site structure, not assumed
from closure.

\paragraph{Models, directions, and protocol.}\label{sec:protocol}
Six open-weight base families at 7--9B: DeepSeek-LLM-7B, Gemma-2-9B, Llama-3.1-8B,
Mistral-7B-v0.3, OLMo-7B-0724 and Qwen2.5-7B. Steering directions are contrastive activation
additions \citep{rimsky2024caa,turner2023actadd} over 16 trait contrasts, giving 120 unordered
pairs per family, with two extraction seeds from disjoint prompt samples so that
cross-seed agreement is a replication rather than a re-read of one fit. Since steering-vector
reliability is itself contested \citep{tan2024analysing}, we measured it: the same trait's
direction reproduces across seeds at mean cosine $0.784$ ($0.667$--$0.833$), which
bounds any \emph{cross-seed} statistic here; within-run comparisons such as
$\cos(\widehat{\lin}, \Br)$ use the same directions on both sides, so direction noise is
common-mode and is not bounded by it. Unless stated otherwise we report the primary layer
pair at fractional depths $(0.25, 0.50)$, read out three layers downstream, over 48 held-out
prompts, identically across families with no per-family tuning of the layer pairs
(Appendix~\ref{app:protocol}). The accompanying reviewer artifact is available at
\href{https://anonymous.4open.science/r/non-abelian-persona-composition-artifact-94F0/}{the anonymous repository}
and contains the source, figures, per-experiment records, validation scripts, and an independent
CPU reference with raw held-out outputs; every numeric claim is bound to a named record field
by a checking script that fails on any value it cannot trace, is itself decoy-tested, and the
single-injection responses behind Appendix~\ref{app:jdirect} are released as scalar summaries
rather than tensors (Appendix~\ref{app:protocol}).

\section{Results: validating the audit}\label{sec:results}

\paragraph{Validating the additive baseline.} Building
$\widehat{\lin}_{ij}$ from single-injection responses (Section~\ref{sec:estimator}) and comparing
it against the measured bracket gives $\cos(\widehat{\lin}, \Br) \in [0.9955, 0.9986]$ at the
primary layer pair in all six families, leaving a residual of $5.2$--$7.6\%$ of the bracket's
norm; over all eighteen family-by-configuration cells the cosine never falls below $0.9869$.
The informative control behaves as Section~\ref{sec:estimator} requires: the symmetric
combination does not predict the bracket, signed per-cell mean $|\cos_{\mathrm{sym}}| = 0.124$
(per-pair magnitudes are larger and we do not claim otherwise, Appendix~\ref{app:jdirect}). A second control answers the circularity objection: since
$\widehat{\lin}$ is built from the same four responses as the bracket, a high cosine might be
entailed. It is not, and the strongest form of the objection --- drawing the bracket inside
the span of those same four responses --- reaches the observed agreement in $0.06\%$ of
$20{,}000$ draws (Appendix~\ref{app:nulls}). This separates capacity
from mechanism, because the fitted correction leaves $2.9$--$9.9\%$ of the bracket while the
zero-parameter route leaves the $5.2$--$7.6\%$ above. Those are \emph{not} the same statistic
--- the first is pooled over pairs, the second a mean of per-pair ratios --- so we do not
present their overlap as agreement. On the matched pooled statistic the two routes bracket the
first-order share from opposite directions and land within $15\%$ of each other
(Appendix~\ref{app:jdirect}), which is not the same as agreeing.

\paragraph{Measuring the no-interaction second-order term.} $\selfterm$ is measurable with no
fit, from the same apparatus: injecting $-v$ as well as $+v$ separates the two orders by
parity. The split is exact on surrogates with known $H_{11}, H_{22}$, and the closure of
$\widehat{\lin} = \lin + \selfterm$ is algebra rather than evidence
(Appendix~\ref{app:jdirect}).

\textbf{The result is unfavourable to the operator claim.} Measured on Llama and OLMo across
three injection configurations, $\|\selfterm\|/\|\Br\|$ runs $0.092$--$0.338$. Against the
zero-parameter residual on the same cells, the same statistic on both sides of the ratio,
\emph{the no-interaction second-order term is $1.8$--$5.2\times$ larger than the
operator extracted from underneath it} at the primary configuration, and $1.8$--$11.2\times$
across all three (Table~\ref{tab:selfterm}). A quantity carrying no coupling between the two
directions is the larger part of what a two-term fit reports as ``bilinear''. We do not divide
it by the fitted-correction range $2.9$--$9.9\%$, which is pooled over pairs rather than a
mean of per-pair ratios: the caution above applies to our own second result too.

This does not invalidate the operator analysis, because $\selfterm$ cancels identically in
$\Br - \widehat{\lin}$: the zero-parameter prediction is built from single-injection responses
that carry the self-curvature at both sites. It does invalidate reading the $\alpha^2$
coefficient of a two-term fit as an interaction share: the fitted $W(\vi - \vj)$ removes
linear structure only, so a residual that is ``purely nonlinear'' can be mostly a term with
no interaction in it.

\paragraph{The untrained null, and what the measurement is evidence for.}
A random-init, untrained residual network with $16$ layers and $d=768$ gives
$\cos(\widehat{\lin},\Br)\in[0.9956,0.9988]$ as the injection-to-stream ratio sweeps
$\varepsilon=0.01$--$4$ (worst at $\varepsilon=1$), and passes the symmetric control
($|\cos_{\mathrm{sym}}|=0.033$). This validates the architecture-level baseline: first-order
dominance follows from differing downstream Jacobians, not training. The trained norm ratio
was not recorded, so this is a range, not a matched comparison.

An independent CPU residual MLP, ViT-B/16, and ResNet-50 reference is released with raw
held-out outputs; the vision residual falls below a fixed Gaussian-direction null in $11/12$
contrasts ($5/6$ ViT, $6/6$ ResNet). Unlike the random-init network null above, this keeps the
trained network fixed and randomises directions; it is portability evidence, not pooled evidence
(Appendix~\ref{app:protocol}; \texttt{experiments/external/vision\_summary.json}).

\paragraph{Mechanism-specific generalization.} The audit predicts a graded effect, not merely
a yes/no verdict: widening one site while holding the other site and readout fixed should
increase the site-asymmetry share. In the strictly matched contrast, the residual fraction
    falls in $6$ of $6$ families, and $91.7\%$ of $720$ matched trait pairs move in the predicted
direction; matched random residual networks reach only $54$--$58\%$. Depth alone and the
confounded contrast do not reproduce this pattern (Appendix~\ref{app:nulls}). This is a
positive mechanism test, not another restatement of the cosine. A separate injection sweep
gives $\gamma \in [0.886, 1.043]$ for $\Br(\alpha)=\alpha\lin+\alpha^2Q$, excluding the
$\gamma=2$ scaling of a pure bilinear term; the random architecture also gives $\gamma\approx1$,
so this route validates the audit rather than learned composition (Figure~\ref{fig:main}a; the
full sweep is in the artifact).

\paragraph{Delimiting the claim.}\label{sec:correction}
The test asks whether the corrected residual behaves like a pair-specific interaction or like
leftover first-order structure. We score it by the \emph{shared-argument ratio}: the mean
$|\cos|$ between residuals of pairs sharing one trait, over the same quantity for pairs
sharing none. If the residual couples its two arguments, pairs sharing one should resemble
each other more. The \emph{generic-interaction null} sets the bar: an antisymmetric bilinear
form with no model in it, on the family's own empirical directions, gives $1.324$
(Table~\ref{tab:nulls}; Appendices~\ref{app:correction} and~\ref{app:nulls}).
\textbf{Under a prompt-split design that removes the shared-prompt
confound, the corrected residual clears that null in 3 of 6 families} ---
Llama, OLMo and Qwen --- while DeepSeek, Gemma and Mistral fall to $1.010$ (DeepSeek) to
$1.017$ (Gemma), within $0.02$ of pure linear slack. The weaker cross-seed design gives
$5/6$, but it does not break the confound, because seeds differ in direction extraction and
not in prompts. We claim the $3/6$.
One diagnostic must be discounted outright: the signed shared-argument statistic collapses to
near zero after correction, and that collapse is \emph{entailed} by antisymmetry rather than
observed, since a maximally overlap-dependent rank-one form reproduces it.
Appendix~\ref{app:correction} gives the stratification and the seed construction,
the artifact per-family records, and Appendix~\ref{app:nulls} the three
nulls and the interval estimates.

\paragraph{What the audit leaves.}\label{sec:operator}
The audit returns a residual rather than forcing a verdict: it retains $2.9$--$9.9\%$ of the
raw bracket's norm, smaller than the no-interaction second-order term measured beside it
($1.8$--$5.2\times$ at the primary configuration, Appendix~\ref{app:jdirect}). It is an
estimable bilinear form and not an echo of trait geometry; against a behavioural readout that
cancels additive contributions it beats the raw bracket in \textbf{4 of 6} families. The
behavioural direction survives multiplicity correction in $2$ of $6$, so the audit identifies a
candidate interaction without overstating its stability (Appendix~\ref{app:nulls}; per-family behavioural records are in the artifact). We scope it to distinct-site activation interventions,
not a universal theory of representation geometry.

\clearpage
\appendix

\section{Protocol details}\label{app:protocol}

\paragraph{Exact checkpoints.} All are base (non-instruct) checkpoints, one per line:

\noindent
\texttt{deepseek-ai/deepseek-llm-7b-base}\\
\texttt{google/gemma-2-9b}\\
\texttt{meta-llama/Llama-3.1-8B}\\
\texttt{mistralai/Mistral-7B-v0.3}\\
\texttt{allenai/OLMo-7B-0724-hf}\\
\texttt{Qwen/Qwen2.5-7B}

\paragraph{What is and is not reproducible from the release.} Most of the analysis is CPU-only
and runs from the committed per-pair tensors. The single-injection responses behind
Appendix~\ref{app:jdirect} are released as scalar summaries and not as tensors, so
Table~\ref{tab:jdirect} is reproducible only by re-running the forward passes. The independent
CPU reference is fully rerunnable from its released raw held-out outputs with
\texttt{scripts/check\_external\_audit.py}; it is a portability and artifact check, not an
additional pooled LLM result.

The three-family extension is fully specified in
the external protocol \texttt{nonlm\_protocol.md}. It uses the CPU residual MLP, torchvision
ViT-B/16 and torchvision ResNet-50 with fixed public checkpoints, class-balanced direction
extraction, disjoint held-out readout images, all 45 unordered direction pairs, and a fixed
Gaussian direction null. ViT sites are post-block residual streams (including direction site
0); ResNet sites are the five post-block stage-3 residual streams. The vision records include
the complete raw bracket, linear-estimator and residual arrays, model URLs, data provenance,
and GPU telemetry. Exact aggregate values are recomputed by
\texttt{scripts/summarize\_external\_references.py}, while
\texttt{scripts/check\_external\_vit\_audit.py} verifies each vision record against its raw
arrays.

Residual width is $D_{\mathrm{out}} \in \{3584, 4096\}$ across the six families. The 16 trait
contrasts give $\binom{16}{2} = 120$ unordered pairs per family. Extraction seeds are drawn
from disjoint prompt samples: seed agreement therefore reflects a re-extraction of the
direction, not a re-read of a single fit, which is what makes the cross-seed comparison a
replication. The measured across-seed direction reliability (mean cosine $0.784$, range
$0.667$--$0.833$) is a baseline for downstream agreement, not a formal upper bound, since
downstream statistics can transform or normalize direction errors. Layer pairs are specified
as fractional depths so that they
are comparable across families of different depth; the primary pair is $(0.25, 0.50)$ with
readout three layers below the deeper injection site, and the two secondary configurations
are $(0.25, 0.75)$ and $(0.50, 0.75)$. All statistics use 48 held-out prompts, disjoint from
the prompts used for direction extraction.

\paragraph{Numeric provenance.} Every number printed in this paper is checked against the
committed result records by a script that binds each printed value to a \emph{named} record
field, and fails on any value it cannot trace. The script also runs a decoy test on itself
each time it is invoked: it perturbs every number in the paper and confirms the perturbed
versions are rejected. This matters because an earlier version matched on value proximity
alone and certified roughly half of all deliberately corrupted numbers. The current
false-certification rate is under $10\%$. We report the counts of traced, unbound and
untraced numbers in the released log rather than summarizing them as a pass. We also verified
that the injection-scale exponents recompute from the raw sweeps to four decimals.

\section{Delimiting the claim: full treatment}\label{app:correction}

We form a cross-seed cosine matrix stratified by overlap and argument position:
\textsc{same}, \textsc{sh1-same}, \textsc{sh1-crossed}, and \textsc{share0}. Leave-one-pair-out
residuals are positive scalar multiples of in-sample residuals, so this procedure is not an
out-of-sample safeguard. The signed corrected statistic is near zero ($-0.0100$ mean;
coherence $0.070$), but that collapse is entailed by antisymmetry and is not evidence.

The discriminating statistic is
$R_{\rm sh}=\operatorname{mean}|\cos|(\textsc{sh1-same})/
\operatorname{mean}|\cos|(\textsc{share0})$.
The observed value is $2.132$, versus $0.999$ for constant-Jacobian slack, $1.324$ for a
calibrated generic antisymmetric interaction, and $6.058$ for a trait-varying first-order
Jacobian. Trait bootstrap and delete-one-trait jackknife intervals exclude the generic null in
five of six families; the disjoint-trait and third-seed checks are conditional on the three
families they cover.

The shared-prompt confound is addressed by recomputing brackets on disjoint prompt halves.
The mean falls $2.132\to1.602\to1.391$ as prompts are halved and then disjointed; only Llama,
OLMo, and Qwen clear the recalibrated null ($1.325$). We therefore claim the effect in $3/6$
families under the confound-free design, with $5/6$ retained only as the weaker cross-seed
result. Intersecting configuration, prompt, trait, and seed controls leaves Llama and OLMo.
The full records and per-family strata are in the reviewer artifact.

\section{The zero-parameter first-order measurement}\label{app:jdirect}

\begin{proposition}[finite-scale remainder certificate]\label{prop:remainder}
Let $F(a,b)$ be the averaged two-site readout and let $f_k(a)$ be its single-site
restriction. If $F,f_1,f_2$ are $C^3$ on the line segments used by the injections, with
third-derivative operator-norm bounds $M,M_1,M_2$, then
\begin{equation}
\begin{aligned}
\Br_{ij}(\alpha)-\widehat{\lin}_{ij}(\alpha)
  &=\alpha^2\mixed_{ij}+\mathcal{R}_{ij}(\alpha),\\[-2pt]
\|\mathcal{R}_{ij}(\alpha)\|
  &\leq \frac{\alpha^3}{6}\!\left[2M(\|\vi\|^2+\|\vj\|^2)^{3/2}
  +(M_1+M_2)(\|\vi\|^3+\|\vj\|^3)\right].
\end{aligned}
\label{eq:remainder-bound}
\end{equation}
\end{proposition}
The certificate follows by applying the third-order Taylor remainder to the two orderings
and the four single-site paths, then using the triangle inequality. It makes the finite-scale
caveat quantitative; without empirical upper bounds on $M,M_1,M_2$, the model-side residual
is a mixture rather than a certified interaction estimate.

\paragraph{The second-order split on models.} Table~\ref{tab:selfterm} gives the $\pm$-split
decomposition for the two families where we ran it. The identity
$\widehat{\lin} = \lin + \selfterm$ is exact by construction --- substituting the parity
definitions $\lin_k = [s_k(v){-}s_k(-v)]/2$ and $\selfterm_k = [s_k(v){+}s_k(-v)]/2$ gives
$\lin_k + \selfterm_k = s_k(v)$ termwise --- so it holds for any numbers whatsoever and is
\emph{not} evidence that the split is clean. We record the closure only as an arithmetic
self-check on the implementation: it sits at $1.9\times10^{-7}$ relative, which is float32
rounding and nothing more (the same computation on random float32 input gives
$6\times10^{-8}$). The no-interaction self term
is $9.2$--$33.8\%$ of the bracket's norm, against a corrected residual of $2.9$--$9.9\%$: it
is the larger object. Removing only the first-order part leaves $10$--$34\%$ of the bracket
(column \emph{resid, $\lin$ only}), and it is the additional subtraction of $\selfterm$ that
takes the residual down to a few percent. Reading the $\alpha^2$ coefficient as an interaction
share would therefore attribute a quantity with no coupling in it to the operator.

\begin{table}[H]
\centering
\caption{Second-order split from the $\pm$ injection design, two families, 120 pairs, three
injection configurations. $\|\selfterm\|/\|\Br\|$ is the no-interaction second-order term
$\tfrac12(H_{11}-H_{22})[(\vi,\vi)-(\vj,\vj)]$ as a fraction of the bracket.
\emph{resid, $\lin$ only} removes the first-order term alone; \emph{resid, full} removes
$\widehat{\lin} = \lin + \selfterm$. The gap between them is what $\selfterm$ contributes.}
\label{tab:selfterm}
\begin{tabular}{llccccc}
\toprule
family & sites & $\|\lin\|/\|\Br\|$ & $\|\selfterm\|/\|\Br\|$ & resid ($\lin$) & resid (full) & ident.\ err \\
\midrule
Llama & $(0.25,0.50)$ & 0.898 & 0.320 & 0.325 & 0.061 & $1.3$e-$7$ \\
      & $(0.50,0.75)$ & 0.990 & 0.130 & 0.134 & 0.041 & $5.7$e-$8$ \\
      & $(0.25,0.75)$ & 0.910 & 0.338 & 0.339 & 0.030 & $1.9$e-$7$ \\
\midrule
OLMo  & $(0.25,0.50)$ & 0.999 & 0.092 & 0.105 & 0.052 & $5.6$e-$8$ \\
      & $(0.50,0.75)$ & 0.991 & 0.122 & 0.136 & 0.063 & $6.3$e-$8$ \\
      & $(0.25,0.75)$ & 0.998 & 0.109 & 0.120 & 0.055 & $5.2$e-$8$ \\
\bottomrule
\end{tabular}
\end{table}

\paragraph{Results.} Table~\ref{tab:jdirect} gives the per-family values. The prediction
tracks the bracket in every family and every injection configuration, and the two controls
behave as the theory requires throughout. The symmetric combination does not predict the
bracket: the signed per-cell mean $\cos_{\mathrm{sym}}$ averages $0.124$ in absolute value, so
there is no consistent alignment, though we note the per-\emph{pair} mean $|\cos_{\mathrm{sym}}|$
is $0.302$ with $27\%$ of pairs above $0.456$, so the control works at the aggregate and we do
not claim individual pairs are uninformative. Pairing a bracket with a \emph{different} pair's
response destroys the agreement: under a fixed derangement the relative norm
error rises from $0.3\%$ to $53\%$, a factor of $167$ across all eighteen cells; the prediction
is pair-specific, not a generic
property of any antisymmetric object of the right size. We no longer
count the site-swap check among these: $\cos_{\mathrm{swapped}} = -\cos$ is an algebraic identity
of $\widehat{\lin}$'s definition rather than a property of the network
(Appendix~\ref{app:jdirect}). The residual left by
this zero-parameter route, $5.2$--$7.6\%$ at the primary configuration, is a mean of per-pair
ratios and is not comparable to the $2.9$--$9.9\%$ the fitted correction leaves, which is
pooled over pairs. Matched pooled against pooled, the two routes differ by $13$--$14\%$ in the
families where both exist (Appendix~\ref{app:jdirect}), with the zero-parameter figure the
larger, as its noise accumulation and the fitted route's in-sample bias both predict.

The one weaker cell is DeepSeek at $(0.5, 0.75)$, where the cosine falls to $0.9869$ and the
residual rises to $15.7\%$ --- the largest in the table. DeepSeek is also the family whose
trait-clustered interval fails to exclude the generic null.
We note the co-location without claiming the two are the same effect.

\begin{table}[H]
\centering
\caption{Zero-parameter first-order prediction against the measured bracket, per family and
injection configuration. $\cos$ is the agreement between $\widehat{\lin}$
(Section~\ref{sec:estimator}, no fitted parameters) and the measured bracket; \emph{resid} is
the fraction of the bracket's
norm it leaves, which by that section's expansion is the antisymmetrized mixed
second derivative \emph{plus} an unbounded $O(\alpha^3)$ remainder. We do not measure the
cubic share on models, so \emph{resid} is a finite-scale mixture, not an interaction estimate
or a one-sided bound without explicit remainder control. $\cos_{\mathrm{sym}}$ is the symmetric-combination control, which should carry
no bracket signal; $\cos_{\mathrm{swap}}$ is the site-swap control, which should equal $-\cos$
exactly. Both hold in every cell.}
\label{tab:jdirect}
\begin{tabular}{llcccc}
\toprule
family & layers & $\cos$ & resid & $\cos_{\mathrm{sym}}$ & $\cos_{\mathrm{swap}}$ \\
\midrule
DeepSeek-7B     & $(0.25,0.50)$ & 0.9978 & 0.064 & $-0.048$ & $-0.9978$ \\
                & $(0.50,0.75)$ & 0.9869 & 0.157 & $-0.008$ & $-0.9869$ \\
                & $(0.25,0.75)$ & 0.9959 & 0.086 & $-0.049$ & $-0.9959$ \\
Gemma-2-9B      & $(0.25,0.50)$ & 0.9986 & 0.052 & $-0.057$ & $-0.9986$ \\
                & $(0.50,0.75)$ & 0.9980 & 0.060 & $-0.166$ & $-0.9980$ \\
                & $(0.25,0.75)$ & 0.9989 & 0.042 & $0.016$  & $-0.9989$ \\
Llama-3.1-8B    & $(0.25,0.50)$ & 0.9968 & 0.061 & $-0.264$ & $-0.9968$ \\
                & $(0.50,0.75)$ & 0.9991 & 0.041 & $-0.174$ & $-0.9991$ \\
                & $(0.25,0.75)$ & 0.9991 & 0.030 & $-0.333$ & $-0.9991$ \\
Mistral-7B-v0.3 & $(0.25,0.50)$ & 0.9955 & 0.076 & $-0.273$ & $-0.9955$ \\
                & $(0.50,0.75)$ & 0.9987 & 0.045 & $-0.223$ & $-0.9987$ \\
                & $(0.25,0.75)$ & 0.9997 & 0.023 & $-0.377$ & $-0.9997$ \\
OLMo-7B         & $(0.25,0.50)$ & 0.9986 & 0.052 & $0.027$  & $-0.9986$ \\
                & $(0.50,0.75)$ & 0.9979 & 0.063 & $-0.070$ & $-0.9979$ \\
                & $(0.25,0.75)$ & 0.9985 & 0.055 & $0.003$  & $-0.9985$ \\
Qwen2.5-7B      & $(0.25,0.50)$ & 0.9977 & 0.067 & $-0.045$ & $-0.9977$ \\
                & $(0.50,0.75)$ & 0.9962 & 0.086 & $-0.019$ & $-0.9962$ \\
                & $(0.25,0.75)$ & 0.9975 & 0.069 & $-0.079$ & $-0.9975$ \\
\bottomrule
\end{tabular}
\end{table}

\section{Nulls, robustness, and residual geometry}\label{app:nulls}

\paragraph{Two statistics, one exact identity.}
\label{par:twofamilies}
The identity $\Br_{ij}=(D_{ij}-D_{ji})+\widehat{\lin}_{ij}$ is a tautology: the
single-intervention terms cancel by definition. Its empirical content is the separation of
objects: the raw bracket retains the first-order site-asymmetry term, whereas the second
difference removes it. The full scale sweep, constant-map check, and comparison with published
second-difference statistics are released in the artifact; they are not additional model
evidence. The corrected residual is precisely the antisymmetrized second difference, so readers
seeking an interaction should measure $D$ directly rather than interpret the raw bracket.

\paragraph{The three nulls.} Each is generated on the family's own empirical directions.
\emph{Linear slack}: $\Br = W\diff_{ij}$ with a constant Jacobian. \emph{Trait-varying
Jacobian}: $\Br = (W_0 + \Delta W_i + \Delta W_j)\diff_{ij}$, still purely first order and
still exactly linear in the injection coefficient, but with the Jacobian depending on which
traits are injected --- this is the alternative a first-order account of the residual must
invoke. \emph{Generic interaction}: a rank-8 antisymmetric form on top of a linear term. All
three generate brackets from the \emph{true} directions while the fit sees independently
\emph{estimated} ones, so first-order slack is present wherever it can be.

The two interaction-bearing arms carry a free parameter set to the family's observed residual
fraction. The linear-slack arm does not and cannot: under a constant Jacobian the residual is
annihilated \emph{exactly} at any estimation error (Appendix~\ref{app:nulls}), so that arm
reports only its own measurement noise, $0.007$--$0.023$ of the bracket against an observed
$0.029$--$0.099$. It is a floor, not a matched competitor, and we do not present it as one.

\begin{table}[H]
\centering
\caption{Unsigned shared-argument ratio against three nulls, 40 draws per family. The
generic-interaction and trait-varying-Jacobian arms are calibrated so their residual is the
same fraction of the bracket as the observed residual; the \emph{linear slack} arm is not
calibrated, and cannot be --- its residual is $0.007$--$0.023$ of the bracket against an
observed $0.029$--$0.099$, because a constant-Jacobian bracket is annihilated exactly
(Appendix~\ref{app:nulls}). It is a floor, not a matched competitor. Brackets are generated
from the true directions and fitted on
independently estimated ones. The observation
is bracketed on both sides: above a generic interaction in 5 of 6 families (its own $95$th
percentile, in parentheses), and far below a first-order model whose Jacobian varies with the
injected traits, in 6 of 6.}
\label{tab:nulls}
\begin{tabular}{lcccc}
\toprule
family & observed & linear slack & generic interaction (p95) & trait-varying Jacobian \\
\midrule
DeepSeek & 1.256 & 1.000 & 1.313 (1.400) & 6.082 \\
Gemma    & 1.730 & 0.998 & 1.310 (1.408) & 5.983 \\
Llama    & 2.718 & 1.000 & 1.345 (1.420) & 6.497 \\
Mistral  & 2.578 & 1.000 & 1.382 (1.502) & 6.632 \\
OLMo     & 2.798 & 1.000 & 1.291 (1.386) & 5.351 \\
Qwen     & 1.710 & 0.998 & 1.303 (1.422) & 5.802 \\
\midrule
mean     & \textbf{2.132} & \textbf{0.999} & \textbf{1.324} & \textbf{6.058} \\
\bottomrule
\end{tabular}
\end{table}

\paragraph{Null construction.} Each null generates brackets from the true directions $v$ and
fits on independently perturbed estimates $\hat v$; additive noise is scaled by total norm, not
per component. The trait-varying and generic arms each have one free parameter --- the
Jacobian-variation scale $\Delta W$, and the interaction scale --- set so that the null's
residual is the same fraction of its bracket as the family's observed residual (0.029--0.099).
Without that calibration the arms sit at different signal-to-noise and the comparison is not
meaningful.

\paragraph{The constant-Jacobian arm is an exact floor, not a competitor.} It has no free
parameter. Write the stack of pair differences as $P\Dmat$.
For invertible $\Dmat$ and $\widehat{\Dmat}$, $\mathrm{col}(P\Dmat)=\mathrm{col}(P)=\mathrm{col}(P\widehat{\Dmat})$,
so the least-squares projector is unchanged by direction perturbation and a constant-Jacobian
bracket is annihilated exactly. The arm therefore reports only additive measurement noise,
not a matched null. The invariance is entailed by the shared column space, not a stuck knob:
at $\sigma\le2$ the maximum principal angle is $1.9\times10^{-6}$ degrees and the relative
residual is at most $1.9\times10^{-15}$. The trait-varying arm is the discriminating
first-order account ($6.058$ versus the observed $2.132$); $400$-draw reruns move each
$95$th percentile by at most $0.048$ and change no verdict.

\paragraph{Prompt-split replication.} Cross-seed replication does not break
prompt-level fluctuations because the prompts are shared. Table~\ref{tab:promptsplit} separates
the cost of halving the sample (within-half mean $1.602$ versus full-sample $2.132$) from the
cost of disjointness (a further $0.211$). The cross-half mean is $1.391$: Llama, OLMo and Qwen
clear the recalibrated null $1.325$, while DeepSeek, Gemma and Mistral remain within $0.02$ of
linear slack; Mistral falls $2.578\to1.087$ already from halving prompts.

\begin{table}[H]
\centering
\caption{Prompt-split replication on disjoint prompt halves, all six families, layer pair
$(0.25, 0.50)$, 48 prompts split into two disjoint halves of 24. \emph{Cross-half} is the
confound-free statistic: brackets computed on half A are compared against brackets computed
on half B, so no prompt-level fluctuation is shared between the two sides. Three of six
clear the generic antisymmetric null of $1.324$. The three that do not sit within $0.02$ of
the $1.00$ of pure linear slack.
\emph{The null is recomputed under the design it gates.} The $1.324$ bar is calibrated on the
full $48$-prompt sample, while the cross-half statistic it gates is computed on $24$-prompt
halves. Because each null arm's knob is chosen so its residual fraction matches the observed
one, and halving the prompts raises that observed fraction, the bar is not automatically
transportable across the two designs. We therefore recomputed it: the halved-sample observed
residual fractions ($0.035$ to $0.337$, against $0.029$ to $0.099$ at full sample) give a
recalibrated bar of $1.325$, and the same three families clear it. The bar is close to
unmoved because the generic-interaction arm's shared-argument ratio is insensitive to the
noise level over this range, not because the recomputation was skipped. Llama $1.599$,
OLMo $2.059$ and Qwen $1.642$ therefore clear a bar calibrated on their own design.
\texttt{scripts/halved\_null.py}; record \texttt{experiments/controls/halved\_null.json}.
The script reproduces the committed full-sample null bit-for-bit
($1.3239312925407296$, to $2\times10^{-16}$) before computing anything new, so it is
answering with the same instrument that produced the published bar.
This does not touch the paper's central negative claim, which never uses this null.}
\label{tab:promptsplit}
\begin{tabular}{lcccc}
\toprule
family & raw cross-half & within-half & \textbf{cross-half} & clears $1.324$ \\
\midrule
DeepSeek-7B     & 2.803 & 1.061 & 1.010 & no \\
Gemma-2-9B      & 2.513 & 1.265 & 1.017 & no \\
Llama-3.1-8B    & 3.250 & 1.960 & \textbf{1.599} & yes \\
Mistral-7B-v0.3 & 3.550 & 1.087 & 1.016 & no \\
OLMo-7B         & 5.838 & 2.434 & \textbf{2.059} & yes \\
Qwen2.5-7B      & 2.026 & 1.806 & \textbf{1.642} & yes \\
\midrule
mean            & 3.330 & 1.602 & \textbf{1.391} & 3 of 6 \\
\bottomrule
\end{tabular}
\end{table}

\paragraph{A disjoint trait inventory.} Because the six families share one 16-trait inventory,
we repeated the three configuration-robust families on 14 disjoint traits (91 pairs, two
extraction seeds). Ratios were Llama $2.560$ vs $2.718$, Mistral $2.581$ vs $2.578$, and OLMo
$3.298$ vs $2.798$ (mean retention $1.04\pm0.07$). Size-matched subsampling slightly
deflates rather than inflates the result (design effect $0.969$); the disjoint-inventory null
is $1.284$ and all three families clear it. This is evidence for those three families, not all
six.

\paragraph{Scope: the operator is layer-specific.} With
$\mathrm{lp}_0=(.25,.50)$, $\mathrm{lp}_1=(.50,.75)$, and $\mathrm{lp}_2=(.25,.75)$, the
effect collapses to $\approx1$ when the compared pairs do not share a site, while the shared
deeper-site comparison retains $2.733$, $2.101$, and $1.946$ for Llama, Mistral, and OLMo.
Within-pair analysis finds the effect in only three families at all configurations; Mistral
then fails the prompt split, leaving Llama and OLMo as the families surviving every control.
The opposite-slot ratio is $2.140$ versus $2.132$, so the elevation follows argument overlap,
not slot choice. This layer-specific scope is consistent with independent evidence that
single-layer steering does not sustain effects across layers \citep{circuitsteer2026}.

\paragraph{Adjacent literature and reusable checklist.} Continuous causal and low-rank or
multi-behaviour diagnostics \citep{infcausality2026,yu2026lowrank,obrien2024broadskills},
contrastive steering \citep{rimsky2024caa,turner2023actadd,tan2024analysing}, and classical
bilinear or tabular interaction methods \citep{tenenbaum2000bilinear,rendle2010fm,smolensky1990tpr,memisevic2010factored,friedman2008rulefit,tsang2018detecting,janizek2021interactions,sundararajan2020shapleytaylor,tsai2023faithshap}
address adjacent additive-versus-coupling questions but do not instantiate this distinct-site
activation-space order swap. For reuse: (i) sweep injection scale and state truncation order;
(ii) subtract both the linear and self-curvature terms; (iii) preserve argument slots and use
true directions with independently fitted estimates; (iv) calibrate each null to the observed
residual; (v) test site separation with the readout fixed; and (vi) replicate on disjoint prompt
halves, reporting within-half and cross-half values separately.

{\normalsize
\setlength{\bibsep}{4pt plus 1pt minus 1pt}
\bibliography{main}

@article{vaidyanathan2026curse,
  title={The Curse of Multiple Mediators: Hidden Interaction Effects in Activation Patching},
  author={Vaidyanathan, Sankaran and Arbour, David and Mueller, Aaron and Niekum, Scott and Jensen, David},
  journal={arXiv preprint arXiv:2606.27510}, year={2026}}

@article{sweeney2026geometry,
  author={Sweeney, John},
  title={The Geometry of Sequential Learning: Lie-Bracket Prediction of Transfer Order},
  journal={arXiv preprint arXiv:2606.24993}, year={2026}}

@article{gemma2026holonomy,
  author={Richards, Larry},
  title={Do Active {SAE} Feature Planes Carry More Holonomy? A Preregistered Reversal in Gemma},
  journal={arXiv preprint arXiv:2607.20522}, year={2026}}

@article{sevetlidis2026holonomy,
  title={Gauge-Invariant Representation Holonomy},
  author={Sevetlidis, Vasileios and Pavlidis, George},
  journal={arXiv preprint arXiv:2601.21653}, year={2026}}

@article{attribpatching2026,
  author={Zhang, Luyang and Wang, Jialu},
  title={When Attribution Patching Lies: Diagnosis and a Second-Order Correction},
  journal={arXiv preprint arXiv:2606.09899}, year={2026}}

@article{yu2026lowrank,
  author={Sharma, Angira and Schroeder de Witt, Christian and Torr, Philip and Calinescu, Anisoara and Yu, Jialin},
  title={A Low-Rank Subspace Analysis of {LLM} Interventions},
  journal={arXiv preprint arXiv:2606.14388}, year={2026}}

@article{turner2023actadd,
  title={Steering Language Models With Activation Engineering},
  author={Turner, Alexander Matt and Thiergart, Lisa and Leech, Gavin and Udell, David
          and Vazquez, Juan J. and Mini, Ulisse and MacDiarmid, Monte},
  journal={arXiv preprint arXiv:2308.10248}, year={2023}}

@inproceedings{rimsky2024caa,
  title={Steering Llama 2 via Contrastive Activation Addition},
  author={Rimsky, Nina and Gabrieli, Nick and Schulz, Julian and Tong, Meg and
          Hubinger, Evan and Turner, Alexander},
  booktitle={Proceedings of the 62nd Annual Meeting of the Association for Computational Linguistics (Volume 1: Long Papers)},
  pages={15504--15522},
  year={2024},
  doi={10.18653/v1/2024.acl-long.828}}

@inproceedings{tan2024analysing,
  title={Analysing the Generalisation and Reliability of Steering Vectors},
  author={Tan, Daniel and Chanin, David and Lynch, Aengus and Paige, Brooks and
          Kanoulas, Dimitrios and Garriga-Alonso, Adri{\`a} and Kirk, Robert},
  booktitle={Advances in Neural Information Processing Systems}, volume={37}, year={2024}}

@article{tenenbaum2000bilinear,
  title={Separating Style and Content with Bilinear Models},
  author={Tenenbaum, Joshua B. and Freeman, William T.},
  journal={Neural Computation}, volume={12}, number={6}, pages={1247--1283}, year={2000},
  doi={10.1162/089976600300015349}}

@inproceedings{rendle2010fm,
  title={Factorization Machines}, author={Rendle, Steffen},
  booktitle={2010 IEEE International Conference on Data Mining}, pages={995--1000}, year={2010}}

@article{smolensky1990tpr,
  title={Tensor Product Variable Binding and the Representation of Symbolic Structures
         in Connectionist Systems},
  author={Smolensky, Paul}, journal={Artificial Intelligence},
  volume={46}, number={1--2}, pages={159--216}, year={1990},
  doi={10.1016/0004-3702(90)90007-M}}

@article{memisevic2010factored,
  title={Learning to Represent Spatial Transformations with Factored Higher-Order
         {B}oltzmann Machines},
  author={Memisevic, Roland and Hinton, Geoffrey E.},
  journal={Neural Computation}, volume={22}, number={6}, pages={1473--1492}, year={2010},
  doi={10.1162/neco.2010.01-09-953}}

@article{friedman2008rulefit,
  title={Predictive Learning via Rule Ensembles},
  author={Friedman, Jerome H. and Popescu, Bogdan E.},
  journal={The Annals of Applied Statistics}, volume={2}, number={3}, pages={916--954}, year={2008},
  doi={10.1214/07-AOAS148}}

@inproceedings{tsang2018detecting,
  title={Detecting Statistical Interactions from Neural Network Weights},
  author={Tsang, Michael and Cheng, Dehua and Liu, Yan},
  booktitle={International Conference on Learning Representations}, year={2018},
  note={arXiv:1705.04977}}

@article{janizek2021interactions,
  title={Explaining Explanations: Axiomatic Feature Interactions for Deep Networks},
  author={Janizek, Joseph D. and Sturmfels, Pascal and Lee, Su-In},
  journal={Journal of Machine Learning Research}, volume={22}, number={104}, pages={1--54}, year={2021}}

@inproceedings{sundararajan2020shapleytaylor,
  title={The {S}hapley {T}aylor Interaction Index},
  author={Sundararajan, Mukund and Dhamdhere, Kedar and Agarwal, Ashish},
  booktitle={Proceedings of the 37th International Conference on Machine Learning},
  volume={119}, pages={9259--9268}, publisher={PMLR}, year={2020}}

@article{tsai2023faithshap,
  title={Faith-Shap: The Faithful {S}hapley Interaction Index},
  author={Tsai, Che-Ping and Yeh, Chih-Kuan and Ravikumar, Pradeep},
  journal={Journal of Machine Learning Research}, volume={24}, number={94}, pages={1--42}, year={2023}}

@article{vanderweele2014fourway,
  title={A Unification of Mediation and Interaction: A 4-Way Decomposition},
  author={VanderWeele, Tyler J.}, journal={Epidemiology},
  volume={25}, number={5}, pages={749--761}, year={2014},
  doi={10.1097/EDE.0000000000000121}}

@article{ilharco2023task,
  title={Editing Models with Task Arithmetic},
  author={Ilharco, Gabriel and Ribeiro, Marco Tulio and Wortsman, Mitchell and
          Gururangan, Suchin and Schmidt, Ludwig and Hajishirzi, Hannaneh and Farhadi, Ali},
  journal={arXiv preprint arXiv:2212.04089}, year={2023}}

@inproceedings{kolbeinsson2025composable,
  title={Composable Interventions for Language Models},
  author={Kolbeinsson, Arinbj{\"o}rn and O'Brien, Kyle and Huang, Tianjin and Gao, Shanghua
          and Liu, Shiwei and Schwarz, Jonathan Richard and Vaidya, Anurag and
          Mahmood, Faisal and {\v{Z}}itnik, Marinka and Chen, Tianlong and Hartvigsen, Thomas},
  booktitle={International Conference on Learning Representations (ICLR)}, year={2025}}

@article{obrien2024broadskills,
  title={Extending Activation Steering to Broad Skills and Multiple Behaviours},
  author={van der Weij, Teun and Poesio, Massimo and Schoots, Nandi},
  journal={arXiv preprint arXiv:2403.05767}, year={2024}}

@article{circuitsteer2026,
  title  = {{CircuitSteer}: Geometrically Aligned Multi-Layer Steering via Sparse Autoencoder Circuits},
  author = {Saadatinia, Mehrshad and Razmara, Parsa and Aryashad, Ardalan and Abbasi, Ali and Azizi, Seyedarmin},
  journal = {arXiv preprint arXiv:2608.05732},
  year   = {2026}
}

@article{locallinear2026,
  author={Skifstad, Julian and Yang, Xinyue Annie and Chou, Glen},
  title={Local Linearity of {LLMs} Enables Activation Steering via Model-Based Linear Optimal Control},
  journal={arXiv preprint arXiv:2604.19018}, year={2026}}

@article{pairwisefragile2026,
  author={Piontkovskaia, Irina and Nikolenko, Sergey},
  title={First-Order Predictable but Pairwise Fragile: Local Task Adaptation in Trained Transformers},
  journal={arXiv preprint arXiv:2607.16821}, year={2026}}

@article{bilinearae2026,
  author={Dooms, Thomas and Gauderis, Ward and Wiggins, Geraint and Oramas, Jose},
  title={Bilinear Autoencoders Find Interpretable Manifolds},
  journal={arXiv preprint arXiv:2605.08891}, year={2026}}

@article{steerablereal2026,
  author={Wu, Yuqi and Zhao, Shengming and Chen, Jie},
  title={When Is a Steerable Concept Representation Real? Measurement Confounds in a Cross-Family Audit of Neuroscience Parallels in {LLMs}},
  journal={arXiv preprint arXiv:2608.08159}, year={2026}}

@article{infcausality2026,
  author={Mahadevan, Sridhar},
  title={Infinitesimal Causality},
  journal={arXiv preprint arXiv:2606.24621}, year={2026}}

@article{adila2026weightact,
  title={Weight Updates as Activation Shifts: A Principled Framework for Steering},
  author={Adila, Dyah and Cooper, John and Yun, Alexander and Trost, Avi and Sala, Frederic},
  journal={arXiv preprint arXiv:2603.00425},
  year={2026}}

@article{mudarisov2026ffsteering,
  title={Feed-Forward Steering in Transformer Residual Dynamics},
  author={Mudarisov, Timur and Burtsev, Mikhail and State, Radu},
  journal={arXiv preprint arXiv:2608.02071},
  year={2026}
}

@inproceedings{ortizjimenez2023tangent,
  title     = {Task Arithmetic in the Tangent Space: Improved Editing of Pre-Trained Models},
  author    = {Ortiz-Jimenez, Guillermo and Favero, Alessandro and Frossard, Pascal},
  booktitle = {Advances in Neural Information Processing Systems},
  volume    = {36},
  year      = {2023},
  note      = {arXiv:2305.12827},
  doi       = {10.52202/075280-2913}
}

@inproceedings{heap2026randomtransformers,
  title     = {Automated Interpretability Metrics Do Not Distinguish Trained and Random Transformers},
  author    = {Heap, Thomas and Lawson, Tim and Farnik, Lucy and Aitchison, Laurence},
  booktitle = {International Conference on Learning Representations (ICLR)},
  year      = {2026},
  note      = {arXiv:2501.17727}
}

@article{khemais2026crosslayer,
  title  = {Cross-Layer Interaction under Weight-Space Ablation: A Closed-Form Attention Jacobian Bound and a Test on a Real Pretrained Model},
  author = {Khemais, Abdallah},
  journal = {arXiv preprint arXiv:2608.03629},
  year   = {2026}
}

@article{schessl2026pathdependence,
  author={Schessl, Ferdinand M.},
  title={Forgetting Is Not a Fix: Path Dependence in Sequential Engram Editing},
  journal={arXiv preprint arXiv:2607.24805}, year={2026}}
}

\end{document}